\documentclass[10pt,twocolumn,letterpaper]{article}
\usepackage{arxiv_layout}
\usepackage{multirow}
\definecolor{categorygray}{gray}{0.85}
\definecolor{linkblue}{rgb}{0.21,0.49,0.74}
\usepackage[pagebackref,breaklinks,colorlinks,allcolors=linkblue]{hyperref}

\title{\textit{FasterGRN}: Reducing the Refinement Redundancy \\  for Accelerating Generative Refinement Networks}

\author{%
Ziqian Zhang\textsuperscript{1,2}\hspace{0.8em}%
Pengcheng Wang\textsuperscript{1}\hspace{0.8em}%
Yulong Chen\textsuperscript{1,2}\hspace{0.8em}%
Yaxing Wang\textsuperscript{3}\hspace{0.8em}%
Senmao Li\textsuperscript{4}\hspace{0.8em}%
Kai Wang\textsuperscript{1,2,*}\\[4pt]
{\normalsize \textsuperscript{1}City University of Hong Kong (Dongguan) \quad
\textsuperscript{2}City University of Hong Kong\quad
\textsuperscript{3}Jilin University}\\
{\normalsize \textsuperscript{4}Mohamed bin Zayed University of Artificial Intelligence}
}

\begin{document}
\twocolumn[{%
    \renewcommand\twocolumn[1][]{#1}%
    \maketitle

    \begin{center}
        \vspace{-6mm}
        \includegraphics[width=0.996\textwidth]{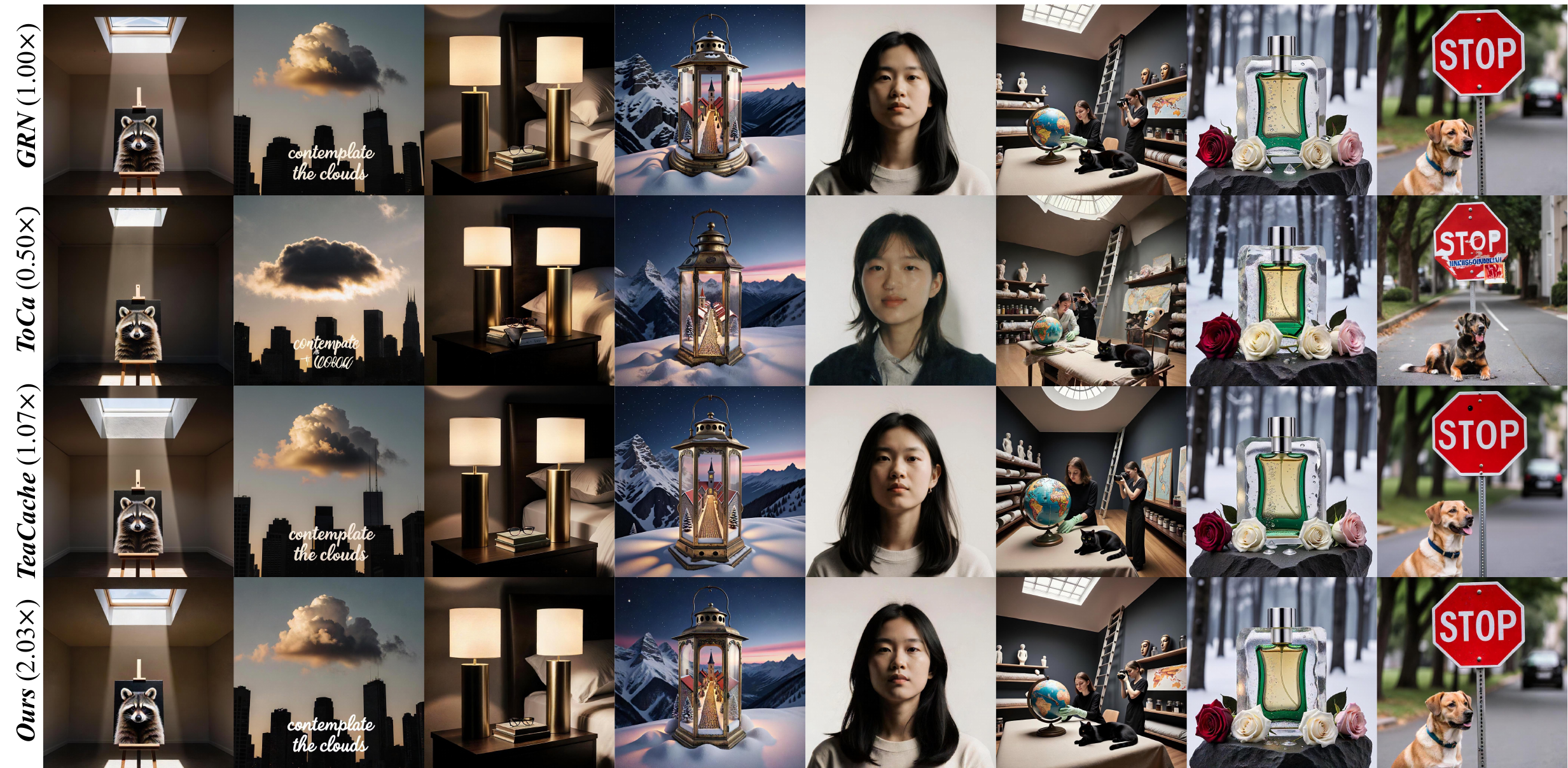}
        \vspace{-6mm}
        \captionof{figure}{
        Image synthesis via applying our \textbf{FasterGRN} and other training-free acceleration methods to the GRN model. By comparison, \textbf{FasterGRN} achieves 2.03$\times$ speedup while outperforming other acceleration approaches and remaining on par with the original GRN. 
        }
        \label{fig:visual}
    \end{center}
}]

\begingroup
\renewcommand{\thefootnote}{*}
\footnotetext[0]{Corresponding author.}
\endgroup

\begin{abstract}
Departing from the paradigm of diffusion and autoregressive models, Generative Refinement Networks (GRN) enables high-quality image generation through iterative refinements.
However, this repeated processing over the entire visual representations results in substantial computation overhead.
Furthermore, the binary representations and global refinement mechanism in GRN disable the direct application of acceleration strategies developed for other generative models.
To accelerate GRN model, we first observe several key properties by analyzing the entire generative process. That includes the stage characteristic, dual branch similarity and more importantly, the \textbf{refinement redundancy}.
Motivated by the observations, we present \textbf{FasterGRN}, a training-free acceleration method for the Generative Refinement Network.
\textbf{FasterGRN} first divides the GRN into the \textit{semantic establishment stage} and \textit{detail refinement stage} according to its semantic establishment. 
For alleviating the \textit{refinement redundancy} mainly resulting from the \textit{detail refinement stage}, we propose \textit{Adaptive Retention Expansion (\textbf{ARE})} and \textit{Selective Token Freezing (\textbf{STF})} to shrink refinement steps and reduce token computations, respectively.
In the experiments on GenEval and DPG-Bench benchmarks, our method \textbf{FasterGRN} achieves a $2.03\times$ sampling speedup while largely preserving image generation performance.
Our evaluation on COCO datasets additionally demonstrates that \textbf{FasterGRN} even outperforms the original GRN in image fidelity.
\end{abstract}

\section{Introduction}
\label{sec:introduction}
Text-to-image generation has attracted considerable attention, with diffusion and autoregressive (AR) models representing two prominent approaches. Diffusion models~\cite{ldm2022,li2024hunyuandit,podell2024sdxl} generate images through iterative denoising, while AR models~\cite{infinity,Parti} progressively predict visual tokens or scales in a prescribed order. 
Building on autoregressive visual modeling, Generative Refinement Networks (GRN)~\cite{han2026grn} introduce a different generation scheme with global refinement mechanism.  GRN iteratively revisits and updates binary visual representations, allowing previously predicted content to be corrected throughout generation. 
Nevertheless, inference efficiency remains a challenge for GRN generation process due to its numerous repetitions of global refinement on the entire visual representations.

Existing acceleration methods for visual generation include training-based~\cite{song2023consistency,salimans2022progressive,yin2024dmd} and training-free approaches~\cite{ma2024deepcache, teacache_2025_CVPR, toca_ICLR2025, taylorseer_2025_ICCV, fastercache2025_518046d8,li2026fastervar}. Training-based methods reduce sampling steps at the cost of additional training, whereas training-free methods exploit inference redundancy without updating model parameters. 
Although training-free methods~\cite{teacache_2025_CVPR,toca_ICLR2025} have achieved substantial speedups in other generative models,
directly applying the existing approaches to GRN generally leads to unsatisfactory results (as shown in Fig.~\ref{fig:visual}).
This motivates us to explore \textit{what properties of GRN can be exploited to reduce redundant computation during generation?}

\begin{figure*}[t]
    \centering
    \includegraphics[width=\textwidth]{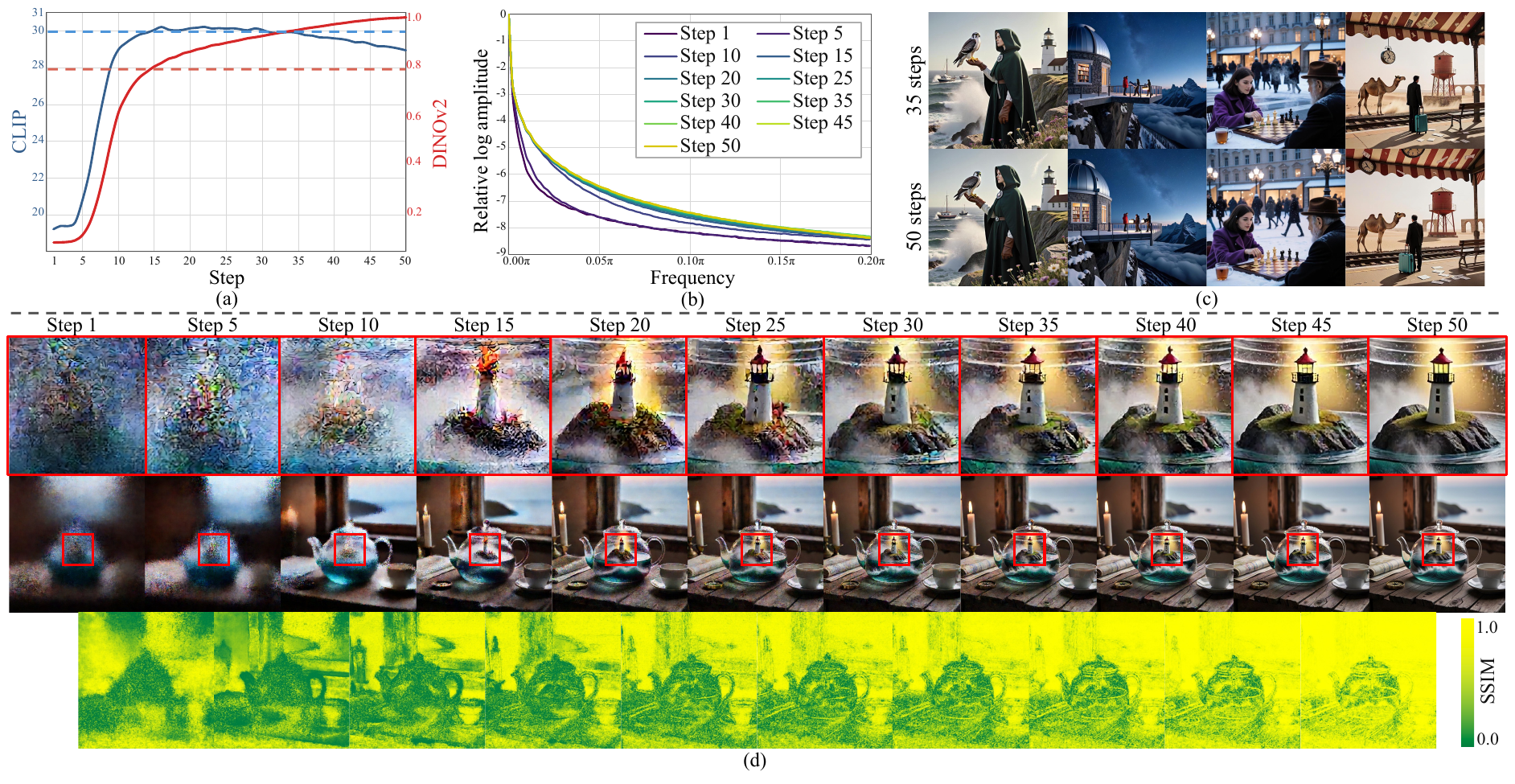}
\vspace{-8mm}
    \caption{
        (a) Visualization of semantic evolution (CLIP and DINOv2)
        across all steps.
        (b) Variations of low frequency features across
        different steps.
        (c) Comparison of samples generated with 35 and 50 refinement steps.
        (d) Visualization of samples across all steps
        (first two rows) and regional differences in the
        magnitude of change across consecutive steps(the third row).
    }
\vspace{-4mm}
    \label{fig:observation}
\end{figure*}

To address this issue, we analyze the text-to-image generation process of GRN model. Inspired by previous works~\cite{yi2024towards,li2026fastervar}, we first examine how semantic content evolves during refinement. 
Our analysis of low-frequency spectra and CLIP~\cite{radford2021clip} similarity suggests that semantic content largely stabilizes in early generation steps (Fig.~\ref{fig:observation}(a,b)).
According to this finding, we separate the GRN generation process as the \textit{semantic establishment stage} followed by the \textit{detail refinement stage}.
{To preserve semantic formation, we retain the original computation during the \textit{semantic establishment stage}, while the relative stability of later predictions suggests opportunities to reduce computation.}
This motivates us to examine the \textit{refinement redundancy} specifically in the \textit{detail refinement stage}.

The first \textit{refinement redundancy} comes from the refinement steps. 
Fig.~\ref{fig:observation}(c) shows that images generated with 35 refinement steps already closely resemble those generated with 50 steps.
Moreover, the top two rows of Fig.~\ref{fig:observation}(d) further show visually similar intermediate predictions across middle and later iterations, suggesting redundancy across refinement steps, which we term \textit{step redundancy}.
To shrink the refinement step while maintaining the generation quality, we propose \textit{Adaptive Retention Expansion (\textit{ARE})}. \textit{ARE} uses the global score to adaptively increase the proportion of predictions retained for the next iteration, reducing the number of refinement steps.
The second \textit{refinement redundancy} results from the token computation. Observing from Fig.~\ref{fig:observation}(d), we suppose that some local regions remain largely stable across refinement steps, as suggested by the SSIM maps, and repeated updates to these regions may therefore be redundant.
To reduce the \textit{token redundancy}, our proposed \textit{Selective Token Freezing (\textit{STF})} uses individual token scores to select locations for freezing, preserving their predictions during noise mixing while allowing continued refinement elsewhere.
Furthermore, Fig.~\ref{fig:cfg} suggests that the unconditional branch can be omitted during the \textit{detail refinement stage} while largely preserving semantic alignment.
The above novel techniques form our \textbf{FasterGRN}, a training-free method that accelerates the GRN model. 

In the experiments, we compare \textbf{FasterGRN} with existing acceleration methods and representative generation models on GenEval~\cite{ghosh2023geneval} and DPG-Bench~\cite{hu2024ella}, and further evaluate generation quality on COCO~\cite{lin2014coco}.
Applied to GRN-2B, our method \textbf{FasterGRN} achieves a $2.03\times$ speedup with small absolute score reductions of 0.008 on GenEval and 0.91 on DPG-Bench.
Meanwhile, it also improves FID~\cite{heusel2017fid}, KID~\cite{binkowski2018kid}, and IS~\cite{salimans2016is} on both COCO 2014 and COCO 2017, which demonstrate that reducing refinement redundancy could even enhance the generation quality.
To summarize, our main contributions are as follows:
\begin{itemize}
    \item We analyze the text-to-image generation process of GRN.  Our analysis reveals the early \textit{semantic establishment stage} and  the semantic stability during the later \textit{detail refinement stage}.  We then identify opportunities to reduce refinement redundancy for GRN acceleration.

    \item We propose \textbf{FasterGRN}, a training-free framework to accelerate GRN. It introduces novel techniques, including \textit{Adaptive Retention Expansion (\textit{ARE})} and \textit{Selective Token Freezing (\textit{STF})}, to reduce refinement iterations and repeated token computation respectively.

    \item We validate \textbf{FasterGRN} through comparisons with existing methods, demonstrating substantial acceleration while largely preserving generation quality. 
    We also demonstrate that \textbf{FasterGRN} can also improve the generation fidelity by reducing refinement redundancy.
\end{itemize}

\section{Related Work}
\label{sec:related}

\paragraph{Text-to-image generation}has evolved through several paradigms. Diffusion and flow models~\cite{dit2023,SDTurbo2024,ldm2022,SD3medium-pmlr-v235-esser24a,sit2024} have demonstrated strong image fidelity and text alignment through iterative generation. Recent autoregressive models~\cite{zero_shot_DALLE,Parti,LlamaGen} show that improved visual tokenizers and model scaling enable discrete next-token prediction to achieve competitive generation quality. 
Visual autoregressive models~\cite{var, infinity} improve generation efficiency through coarse-to-fine next-scale prediction, with bitwise modeling further enhancing discrete representation capacity. Nevertheless, their generation process remains organized around a prescribed scale progression. 
GRN~\cite{han2026grn} introduces a different formulation, combining Hierarchical Binary Quantization with global refinement to allow previously sampled bits to be refined throughout generation. 
Our work builds on this novel generation scheme and gives the first observation to explicitly divide this generation process into \textit{semantic establishment stage} and \textit{detail refinement stage}, which the previous GRN-based works ignored~\cite{chen2026refinedit,xie2026grnedit}.
Our proposed \textbf{FasterGRN} reduces redundant computation while retaining its ability to refine generated content. 

\paragraph{Existing acceleration methods} can be broadly categorized into training-based and training-free approaches. Training-based methods typically employ distillation~\cite{salimans2022progressive,yin2024dmd} or model compression~\cite{Fang2025CVPR_tinyfusion,structurepruning} to reduce sampling steps or per-step computation, at the cost of additional training. 
For diffusion models, training-free approaches mainly explore efficient sampling~\cite{lu2022dpmsolver,zhao2023unipc}, token reduction~\cite{tomesd_2023_CVPR,similarityTokenPruning,cha2026tokensmergediffusiondynamics,justintime_2026_CVPR}, and feature caching or prediction~\cite{ma2024deepcache,fastercache2025_518046d8,teacache_2025_CVPR,toca_ICLR2025,taylorseer_2025_ICCV,silveria2025chipmunktrainingfreeaccelerationdiffusion}, reducing either the number of model evaluations or the cost of each forward pass. 
For autoregressive models, parallel decoding reduces sequential model evaluations. For example, speculative decoding~\cite{speculative_decoding, speculative_jacobi_decoding} generates candidate tokens with a lightweight draft model and verifies them in parallel with the target model. 
Such strategies target at accelerating causal token generation, and cannot be directly applied to the GRN tokens which are globally revised in each step.
For visual autoregressive models, FasterVAR~\cite{li2026fastervar} takes a stage-wise perspective. It distinguishes early semantic and structural establishment from later fidelity refinement, then it exploits redundancy in the later stage. 
Likewise, our analysis shows that semantic alignment tends to stabilize before refinement concludes. Building on this perspective, our \textbf{FasterGRN} combines stage-wise observation with computation reduction in later stage.

\section{Method}
\label{sec:method}
In this section, we first introduce the preliminaries of Generative Refinement Networks (GRN) in Sec.~\ref{sec:preliminary}.
Then we reveal the stage characteristics of GRN generation in Sec.~\ref{observation:two-stage}. Finally in Sec.~\ref{refinement redundancy}, we analyze the redundancy in \textit{detail refinement stage} and propose \textit{Adaptive Retention Expansion (ARE)} and \textit{Selective Token Freezing (STF)}.

\begin{figure*}[t]
    \centering
    \includegraphics[width=\textwidth]{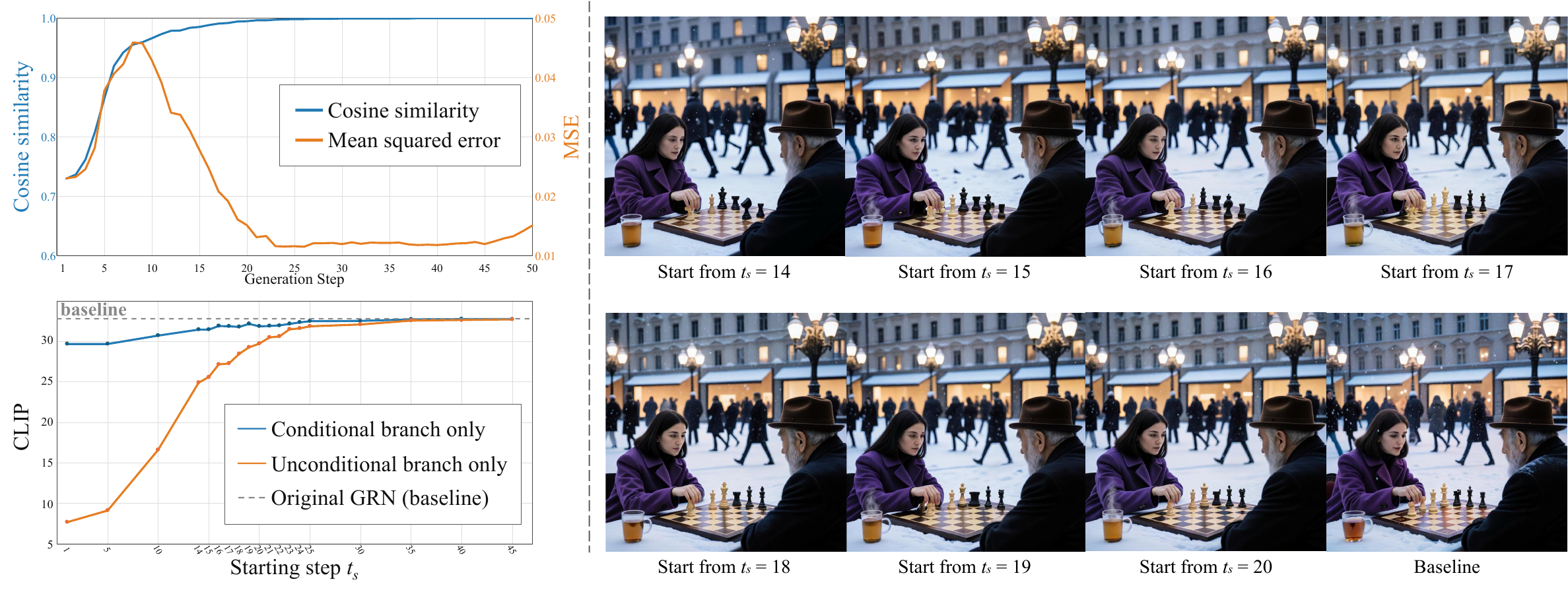}
\vspace{-8mm}
    \caption{
        (Up-Left) Cosine similarity changes
        over steps.
        (Down-Left) Evaluation of semantics when switching
        to the conditional or unconditional branch at
        different steps.
        (Right) Sample visualizations obtained by switching to the conditional branch only at later steps.
    }
\vspace{-4mm}
    \label{fig:cfg}
\end{figure*}

\subsection{Preliminaries}
\label{sec:preliminary}
\paragraph{Binary image representation.}
A frozen Hierarchical Binary Quantization (HBQ) tokenizer $\mathcal{E}_{\mathrm{HBQ}}$ encodes an image $I$ as $Y=\mathcal{E}_{\mathrm{HBQ}}(I)\in\{0,1\}^{CN\times D}$, where $N$ is the number of tokens, $C$ is the number of channels in one token, and $D$ is the number of bits per channel. Different bits contribute distinctively to the reconstruction of continuous values.

\paragraph{Generative refinement.}
GRN initializes $Y_0=Z$, where each bit of random code $Z$ is sampled once as $0$ or $1$ with equal probability. The same $Z$ is reused throughout refinement.
Given the binary input \(Y_t\) at this step \(t\) and text condition, GRN predicts all binary coordinates in parallel, and gives the probabilities $P_t$ of each bit being 1 or 0. Sampling from $\{0, 1\}$ with $P_t$ gives the predicted bits in this step $\widehat{Y}_t$.
Following that, the input to next step combines sampled predictions with the fixed random code:
\begin{equation}
    Y_{t+1}
    =
    S_t\odot\widehat{Y}_t
    +(1-S_t)\odot Z,
    \label{eq:grn_refinement}
\end{equation}
where $\odot$ denotes the element-wise multiplication operator.
The selection mask \(S_t\) satisfies
$S_{t,n,c,d}\sim\operatorname{Bernoulli}(l_{t+1})$:
it equals $1$ with probability $l_{t+1}$ and $0$ otherwise, where $n \le N$, $c \le C$, and $d\le D$ represent the index of token, channel, and bit respectively.
Thus, $l_{t+1}$ controls the expected proportion of
predictions retained in the next state.
And this proportion of retained bits can be represented as an increasing function $L(t)$ mapping from $t\in(0, T]$ to $l_t\in(0, 1)$, which will be used in Sec.~\ref{par:AdaInc}.

\subsection{Stage Characteristics}
\label{observation:two-stage}

To understand the GRN generation process\footnote{The observation is done with an 80-prompt sample sets and the samples in this set are excluded in final evaluation.}, we analyze how GRN predictions evolve by decoding the binary representations into images at successive refinement steps. As the retention ratio $l_t$ increases, the main semantic content becomes recognizable early, while later steps mainly refine textures and fine details.

To examine this phenomenon, we first use CLIP~\cite{radford2021clip} and DINOv2~\cite{dinov2} metrics to measure the global semantic alignment for intermediate generated images. Fig.~\ref{fig:observation}(a) shows that the CLIP score rises rapidly during early generation and stabilizes around steps 15 to 20. Meanwhile, the \textit{low-frequency spectra} in Fig.~\ref{fig:observation}(b) also become close to the final output around step 15. These observations suggest that global semantics and coarse structure are largely established in early generation steps. 
Furthermore, the intermediate images in Fig.~\ref{fig:observation}(d) further show that later updates mainly affect local details. Based on these observations, we characterize generation as a \textit{semantic establishment stage} followed by a \textit{detail refinement stage}.

During the \textit{semantic establishment stage}, the low retention ratio leaves a large proportion of the input from random code, while global semantics and structure continue to evolve rapidly. 
Later \textit{detail refinement stage} retains more predictions and mainly refines local details. 
This distinction motivates focusing acceleration on the \textit{detail refinement stage} while preserving the original process during the \textit{semantic establishment stage}. 

\paragraph{Classifier-Free Guidance Control.}
During the \textit{detail refinement stage}, the first redundant computation we have noticed is the Classifier-Free Guidance (CFG).
Fig.~\ref{fig:cfg} shows that conditional and unconditional predictions become increasingly similar in this stage. Switching to the conditional branch alone around steps 15 to 20 also yields final semantic alignment close to the original GRN. 
Reasonably, we disable CFG during \textit{detail refinement stage} and retain only the conditional branch. 
This operation removes the unnecessary unconditional computation while preserving direct text conditioning. 
This serves as a simple acceleration technique. 
Following that, we further reduce the \textbf{refinement redundancy of GRN} during \textit{detail refinement stage}.

\subsection{Refinement Redundancy of GRN}
\label{refinement redundancy}
\begin{figure*}[t]
    \centering
    \includegraphics[width=\textwidth]{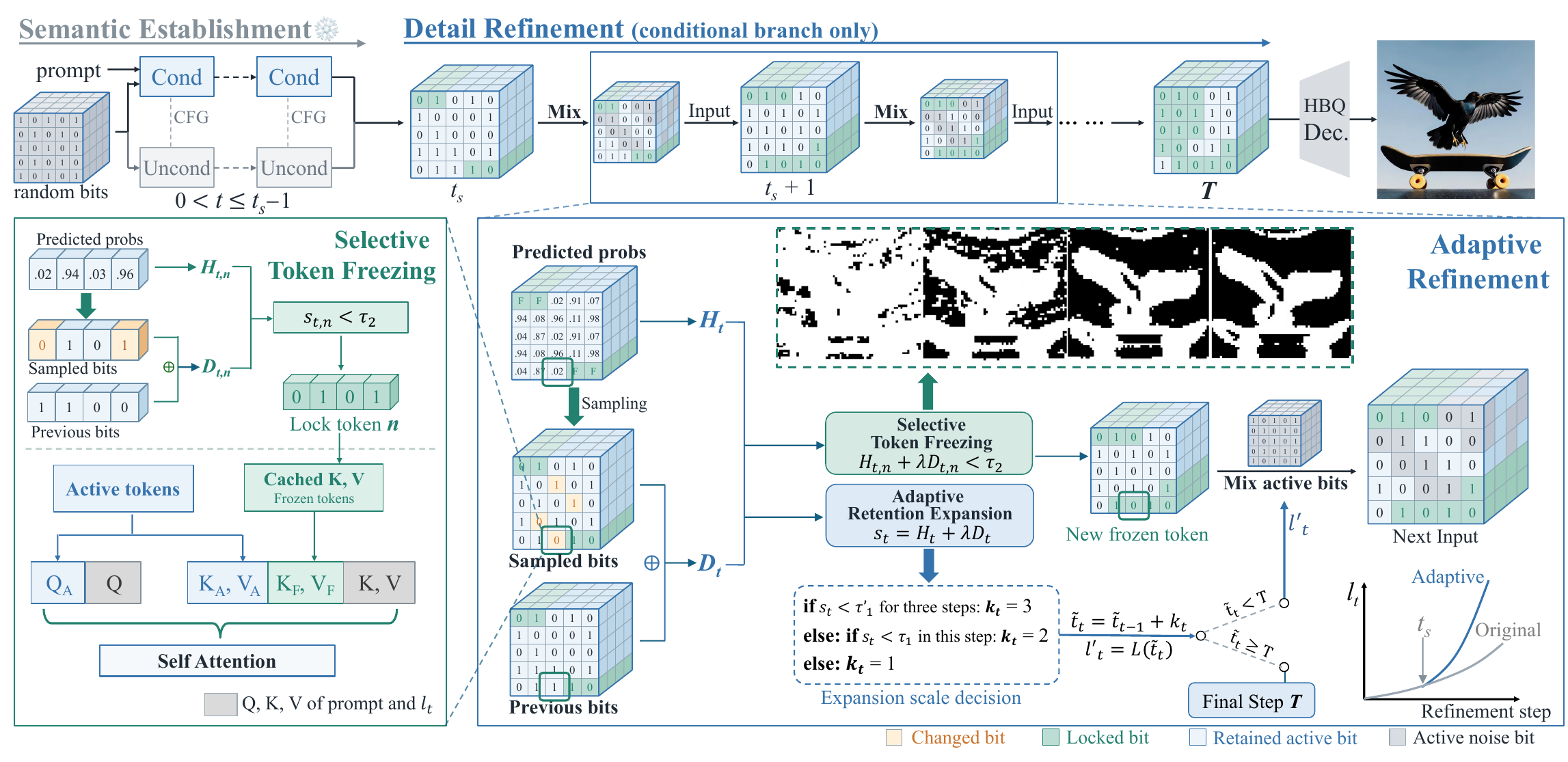}
    \vspace{-8mm}
    \caption{
        Overview of the \textbf{FasterGRN} framework. In the \textit{detail refinement stage}, we activate \textit{Selective Token Freezing} and \textit{Adaptive Retention Expansion} to achieve adaptive refinement.
    }
    \vspace{-5mm}
    \label{fig:framework}
\end{figure*}

In this section, we divide our found refinement redundancy into \textit{sampling steps redundancy} across iterations and \textit{latent tokens redundancy} across spatial locations. 
As shown in Fig.~\ref{fig:framework}, our proposed \textbf{FasterGRN} framework addresses them by \textit{Adaptive Retention Expansion (ARE)} and \textit{Selective Token Freezing (STF)}, respectively. 
Based on the stage characteristics, we use only the conditional branch and activate \textit{ARE} and \textit{STF} mechanisms in \textit{detail refinement stage}.

\subsubsection{Sampling Steps Redundancy}

Sampling steps redundancy arises when the retention schedule requires more refinement iterations than the current predictions need. GRN's complexity-aware sampling uses prediction entropy at the 5th step to select the retention growth rate for the remaining iterations. Although this adapts the schedule across samples, the chosen rate remains fixed and does not respond to later changes in prediction confidence.%

Fig.~\ref{fig:observation}(c) compares final outputs generated with maximum budgets of 35 and 50 steps.
The outputs remain visually similar across the sampled images. The key difference lies in the growth of the retention ratio: the shorter schedule traverses the retention range with fewer updates and thus uses larger average increments of $l_t$. This suggests that a slower increase is not always needed, motivating \textit{adaptive retention expansion} based on the evolving prediction state.

\paragraph{Adaptive Retention Expansion (ARE).}
\phantomsection
\label{par:AdaInc}
\textit{ARE} advances the retention schedule according to prediction confidence and temporal stability. We use predictive entropy to measure prediction uncertainty and the cross-step bit-flip ratio to assess temporal stability. 

Let $p_{t,n,c,d}$ denote the predicted probability that bit $d$ in channel $c$ of token $n$ equals 1 at step $t$, and let $\hat y_{t,n,c,d}$ denote its sampled value. We weight the four bits in each channel by their contributions to reconstruction, using $\alpha_d=2^{4-d}/15$ for $d\in\{1,2,3,4\}$.
The weighted predictive entropy is computed as:
\begin{equation}
H_{t,n}
=
\sum_{d=1}^{4}\alpha_d
\frac{1}{C}\sum_{c=1}^{C}
h(p_{t,n,c,d}),
\label{eq:refinement_entropy}
\end{equation}
where $h(p)=-p\log_2 p-(1-p)\log_2(1-p)$.
The weighted cross-step bit-flip ratio is defined as:
\begin{equation}
D_{t,n}
=
\sum_{d=1}^{4}\alpha_d
\frac{1}{C}\sum_{c=1}^{C}
\mathbb{I}\!\left[
\hat y_{t,n,c,d}\neq\hat y_{t-1,n,c,d}
\right],
\label{eq:refinement_bit_flip}
\end{equation}
where $\mathbb{I}[\cdot]$ is the indicator function. These signals measure prediction uncertainty and variation between successive sampled predictions, respectively.

\textit{ARE} aggregates both signals to obtain a global score:
\begin{equation}
s_t
=
\frac{1}{N}\sum_{n=1}^{N}H_{t,n}
+
\lambda
\frac{1}{N}\sum_{n=1}^{N}D_{t,n},
\end{equation}
where $N$ is the number of tokens and $\lambda$ controls the contribution of the bit-flip ratio relative to predictive entropy in the global score. A lower score indicates confident and stable predictions, supporting faster retention growth. 

Using the original GRN retention schedule $L(\cdot)$, \textit{ARE} advances a virtual step index $\tilde{t}_t$ by an adaptive stride $k_t$:
\begin{equation}
\tilde{t}_t=\min\!\left(T,\tilde{t}_{t-1}+k_t\right),
\qquad
l'_t=L(\tilde{t}_t).
\end{equation}
Here, $t$ counts actual refinement steps, while $\tilde{t}_t$ indexes the original GRN schedule with terminal index $T$. The adaptive stride $k_t$ is determined by:

\begin{equation}
k_t=
\begin{cases}
3, & s_j<\tau'_1,
     \quad \forall j\in\{t-2,t-1,t\},\\
2, & s_t<\tau_1,\\
1, & \text{otherwise}.
\end{cases}
\end{equation}

The conditions are checked from top to bottom, and the first satisfied case determines $k_t$.
\textit{ARE} reduces redundancy at the step level by adaptively expanding the retention schedule.
However, within each remaining step, GRN still repeatedly processes all spatial tokens whose predictions may have already stabilized.
This indicates an opportunity to reduce computation at the token level, and motivates a closer examination of token redundancy during refinement.

\subsubsection{Latent Token Redundancy}

Latent tokens redundancy arises from repeatedly updating spatial locations that have already stabilized. The cross-step SSIM maps in Fig.~\ref{fig:observation}(d) show that some regions remain highly similar during later refinement, while others continue to change. Updating all tokens at every step therefore allocates computation to both stable and evolving regions. This motivates reusing stable tokens by our proposed technique termed \textit{Selective Token Freezing (STF)} while continuing to refine the remaining locations.

\paragraph{Selective Token Freezing (STF).}
\phantomsection
\label{par:FTL}
Our proposed \textit{STF} uses the token-level entropy and bit-flip ratio defined above to identify predictions that are both confident and stable. For each token, we compute $s_{t,n}$ as the token score:
\begin{equation}
s_{t,n}
=
H_{t,n}
+
\lambda D_{t,n},
\end{equation}
where $\lambda$ weights temporal stability and $\tau_2$ controls the confidence and stability required for token freezing. A token is eligible for freezing when $s_{t,n}<\tau_2$. 
\textit{STF} uses token-wise entropy $H_{t,n}$ and the bit-flip ratio $D_{t,n}$ totally following the same definition as in Eq.~\ref{eq:refinement_entropy} and Eq.~\ref{eq:refinement_bit_flip} respectively.
Once a token is frozen, its retained prediction and cached keys and values are reused in subsequent steps. To preserve frozen predictions, noise mixing is restricted to active tokens, and the retention ratio is applied within this active set. Frozen tokens thus remain unchanged, while active tokens continue to be refined.

\medskip

Overall, our proposals including CFG control, \textit{ARE}, and \textit{STF} form the \textbf{FasterGRN} framework, a training-free acceleration method that preserves semantic establishment and speeds up detail refinement for the GRN model.

\section{Experiments}
\label{sec:experiments}
We first describe the experimental setup in Sec.~\ref{sec:experimental_setup}.
We then evaluate the quality and efficiency of \textbf{FasterGRN} for text-to-image generation in Sec.~\ref{sec:main_results}, with additional qualitative examples of video generation.
Finally, we examine the effects of the stage boundary and individual acceleration components through ablation studies in Sec.~\ref{sec:ablation_analysis}.
\subsection{Experimental Setup}
\label{sec:experimental_setup}

\paragraph{Benchmarks.}
We evaluate \textbf{FasterGRN} on three complementary benchmarks: GenEval~\cite{ghosh2023geneval}, DPG-Bench~\cite{hu2024ella}, and COCO~\cite{lin2014coco}.
GenEval uses 553 prompts to test whether generated images depict the requested objects, counts, colors, and spatial relations.
DPG-Bench uses 1,065 detailed prompts to assess whether a model can satisfy multiple semantic constraints within a single image.
For both benchmarks, we generate four images per prompt using seeds 42, 43, 44, and 45.
On COCO2014-30K and COCO2017-5K, we compare GRN and \textbf{FasterGRN} in terms of how closely generated image distributions match those of real COCO images.
Together, these benchmarks allow us to assess gains in inference efficiency alongside changes in prompt adherence and image fidelity relative to the original GRN.

\paragraph{Evaluation Metrics.}
We assess \textbf{FasterGRN} in terms of prompt adherence, image quality, and inference efficiency.
GenEval and DPG-Bench scores measure compositional correctness and adherence to detailed prompts, respectively.
For image quality, we use Fr\'echet Inception Distance (FID)~\cite{heusel2017fid} and Kernel Inception Distance (KID)~\cite{binkowski2018kid} to measure how closely the generated images match the real image distribution in feature space. Inception Score (IS)~\cite{salimans2016is} complements these metrics by assessing object recognizability and category diversity.
In the ablation studies, we use CLIP similarity~\cite{radford2021clip} to measure text--image alignment and ImageReward (IR)~\cite{imagereward} and HPSv2.1~\cite{hpsv2} to estimate human preference.
To quantify the trade-off between quality and efficiency, we report these scores alongside sampling latency and speedup relative to the original GRN.
All experiments are conducted on a single NVIDIA H100 GPU.

\paragraph{Implementation Details.}
Evaluated acceleration methods use a frozen GRN without additional training. We select the starting step $t_s=17$ of \textit{detail refinement stage} based on our observation in Fig.~\ref{fig:cfg}. $\lambda$ is set to be 0.3 to balance the weights of entropy and bit-flip ratio used in the score. Thresholds of scores $\tau_1$, $\tau'_1$, and $\tau_2$ are set to be 0.9, 0.5, and 0.2 respectively. For generation resolutions, LlamaGen-XL and Emu3 retain their native $512\times512$ resolution, while all other models generate $1024\times1024$ images.

\begin{table*}[t]
\centering
\caption{Quantitative comparisons of perceptual quality on the GenEval and DPG-Bench Benchmarks.}
    \vspace{-3mm}
\label{tab:main_results}
\setlength{\tabcolsep}{4pt}
\renewcommand{\arraystretch}{1.08}
\resizebox{\textwidth}{!}{
\begin{tabular}{c|lccc|cccc|cccc}
\toprule
\multirow{2}{*}{Paradigm} &
\multirow{2}{*}{Method} &
\multirow{2}{*}{Latency$\downarrow$} &
\multirow{2}{*}{Speedup$\uparrow$} &
\multirow{2}{*}{Params} &
\multicolumn{4}{c|}{GenEval$\uparrow$} &
\multicolumn{4}{c}{DPG-Bench$\uparrow$} \\
\cmidrule(lr){6-9}
\cmidrule(lr){10-13}
& & & & &
Two obj. & Position & Color Attr. & Overall &
Entity & Attribute & Relation & Overall \\
\midrule

\multirow{2}{*}{Diffusion}
& SDXL~\cite{podell2024sdxl}
& 2.84s & - & 2.6B
& 0.74 & 0.15 & 0.23 & 0.55
& 82.40 & 80.90 & 86.80 & 74.70 \\
& Hunyuan-DiT~\cite{li2024hunyuandit}
& 5.80s & - & 1.5B
& 0.77 & 0.13 & 0.30 & 0.63
& 80.60 & 88.00 & 74.40 & 78.90 \\
\midrule

\multirow{2}{*}{AR}
& LlamaGen-XL~\cite{LlamaGen}
& 24.36s & - & 0.8B
& 0.34 & 0.07 & 0.04 & 0.32
& 75.32 & 76.25 & 86.00 & 65.16 \\
& Emu3~\cite{wang2024emu3}
& 196.30s & - & 8.5B
& 0.71 & 0.17 & 0.21 & 0.54
& 86.68 & 86.84 & 90.22 & 80.60 \\
\midrule

\multirow{2}{*}{VAR}
& Infinity-2B~\cite{infinity}
& 1.01s & - & 2B
& 0.79 & 0.24 & 0.58 & 0.69
& 90.81 & 87.89 & 88.19 & 83.41 \\
& Infinity-8B~\cite{infinity}
& 2.53s & - & 8B
& 0.97 & 0.61 & 0.75 & 0.83
& 90.92 & 88.83 & 93.57 & 86.68 \\
\midrule

\multirow{2}{*}{Flow}
& SD3-Medium~\cite{SD3medium-pmlr-v235-esser24a}
& 2.37s & - & 2B
& 0.74 & 0.34 & 0.36 & 0.62
& 91.00 & 88.80 & 80.70 & 84.10 \\
& FLUX.1-dev~\cite{flux2024}
& 9.30s & - & 12B
& 0.81 & 0.20 & 0.47 & 0.67
& 89.50 & 88.70 & 91.10 & 84.00 \\
\midrule

\multirow{5}{*}{GRN}
& GRN~\cite{han2026grn}
& 6.55s & $1.00\times$ & 2B
& 0.89 & 0.37 & 0.58 & 0.72
& 92.21 & 88.86 & 94.06 & 85.64 \\

& +TaylorSeer~\cite{taylorseer_2025_ICCV}
& 4.40s & $1.49\times$ & 2B
& 0.31 & 0.08 & 0.06 & 0.26
& 79.85 & 74.84 & 88.78 & 67.67 \\

& +ToCa~\cite{toca_ICLR2025}
& 13.21s & $0.50\times$ & 2B
& 0.85 & 0.37 & 0.56 & 0.70
& 92.56 & 89.60 & 94.25 & 86.53 \\

& +TeaCache~\cite{teacache_2025_CVPR}
& 6.12s & $1.07\times$ & 2B
& 0.88 & 0.35 & 0.57 & 0.71
& 92.16 & 88.98 & 93.56 & 85.86 \\

& +\textbf{FasterGRN (Ours)}
& 3.23s & $2.03\times$ & 2B
& 0.89 & 0.34 & 0.57 & 0.71
& 91.34 & 88.23 & 93.44 & 84.73 \\

\bottomrule
\end{tabular}
}
\end{table*}

\begin{figure*}[t]
    \centering
    \vspace{-3mm}
    \includegraphics[width=0.92\linewidth]{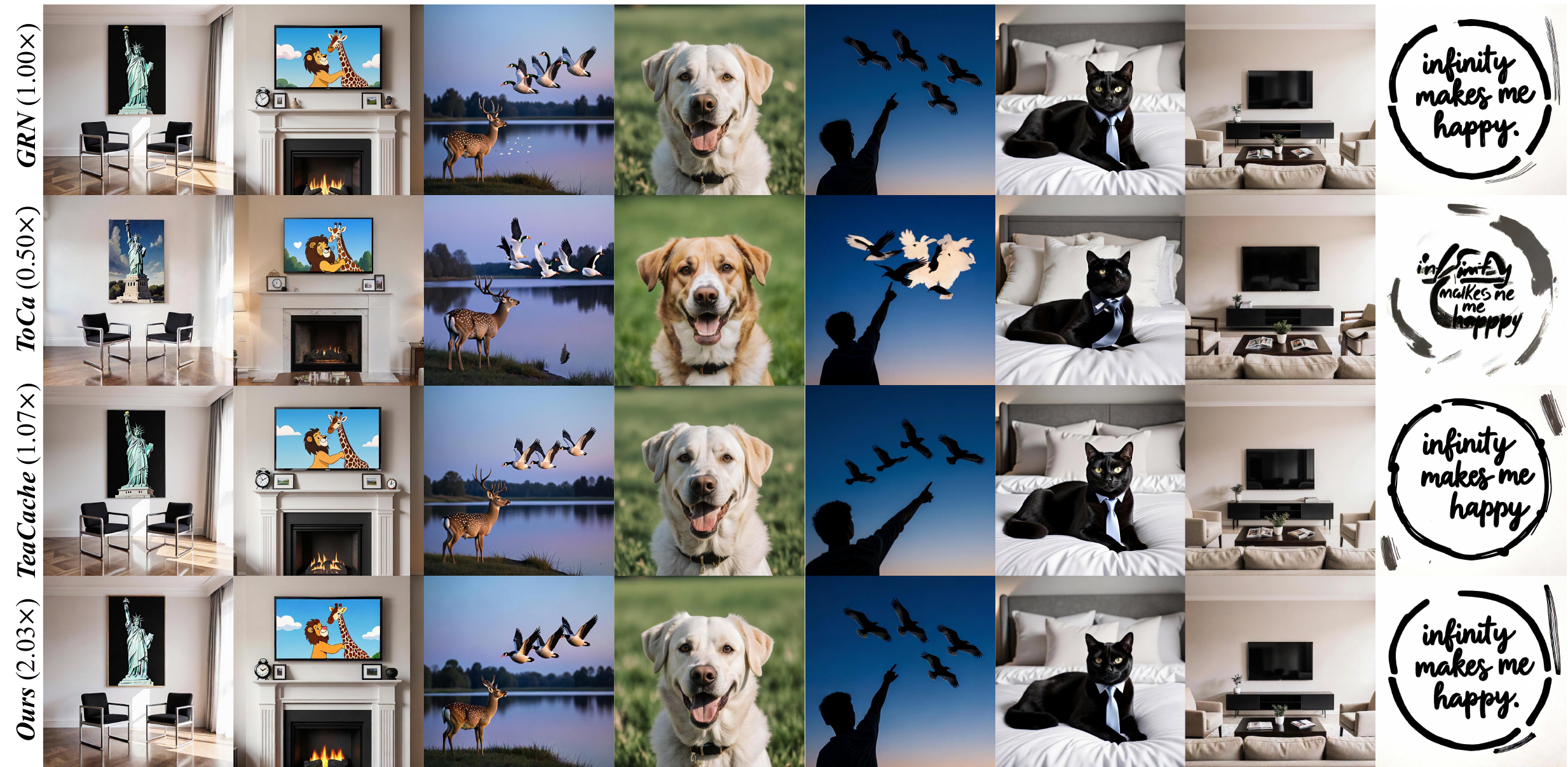}
    \vspace{-1mm}
    \caption{
        More qualitative comparison of image synthesis via applying \textbf{FasterGRN} and other acceleration methods.
    }
    \vspace{-4mm}
    \label{fig:visual_doublecolumn}
\end{figure*}

\subsection{Main Results}
\label{sec:main_results}
\begin{figure}[t]
    \centering
    \vspace{-2mm}
    \includegraphics[width=\linewidth]{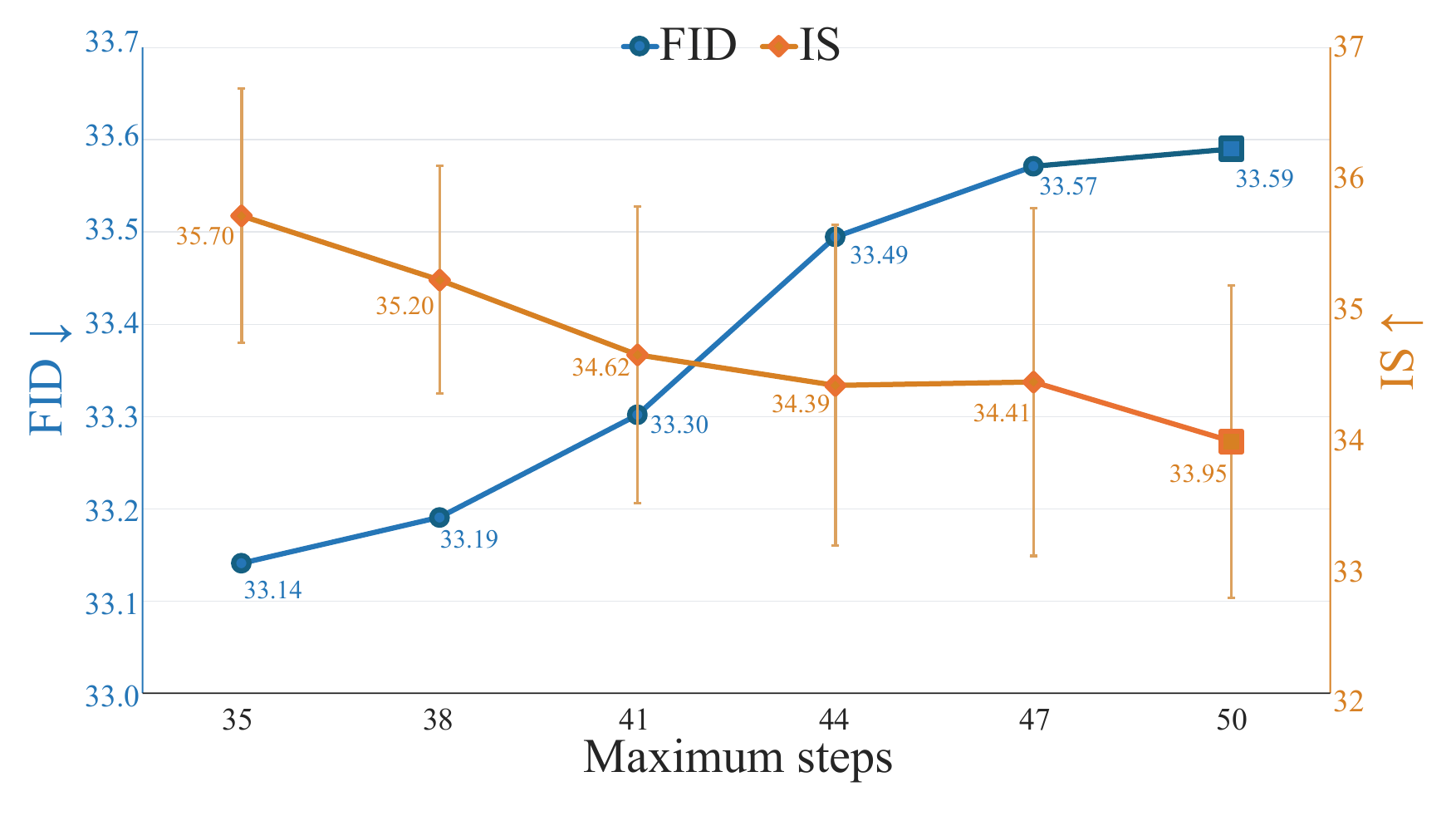}
    \vspace{-8mm}
    \caption{
    Effect of refinement steps on image fidelity.
    }
    \vspace{-5mm}
    \label{fig:FID_IS_change}
\end{figure}

\paragraph{Comparison with Baselines and Other Methods.}
GRN is competitive across diverse generation paradigms, and \textbf{FasterGRN} largely preserves this performance while reducing sampling time.
In Table~\ref{tab:main_results}, GRN-2B ranks second among the evaluated backbones, scoring 0.72 on GenEval and 85.64 on DPG-Bench. It outperforms Infinity-2B at the same parameter size and FLUX.1-dev with six times as many parameters.
Among the acceleration methods applied to GRN, TeaCache~\cite{teacache_2025_CVPR} matches the GenEval score of \textbf{FasterGRN} but offers only a modest speedup. ToCa~\cite{toca_ICLR2025} improves on the original GRN on DPG-Bench but more than doubles sampling time, while TaylorSeer~\cite{taylorseer_2025_ICCV} reduces latency at the cost of a substantial decline in compositional accuracy.
Examples in Fig.~\ref{fig:visual_doublecolumn} show how closely the outputs of \textbf{FasterGRN} resemble those of GRN in both scene composition and local detail. The two chairs retain their positions beneath the Statue of Liberty artwork, and fine fur texture remains visible in the dog portrait.
Overall, \textbf{FasterGRN} delivers the largest sampling speedup among the methods evaluated on GRN, achieving a $2.03\times$ speedup while keeping performances close to the original model.
\vspace{-4mm}
\paragraph{Improved Image Fidelity.}
Despite using less computation, \textbf{FasterGRN} improves image fidelity over the original GRN on both COCO evaluation sets.
In Table~\ref{tab:coco_comparison}, FID decreases from 33.59 to 32.16 on COCO2017-5K, with a similar improvement on COCO2014-30K. KID also decreases and IS increases on both sets.
To examine the role of later refinement steps, we keep the early refinement schedule fixed and vary the total sampling budget on COCO2017-5K.
Fig.~\ref{fig:FID_IS_change} shows that reducing the budget from 50 to 35 steps lowers FID to 33.14 and improves IS, suggesting that additional detail refinement does not necessarily improve fidelity.
Together, these results show that \textbf{FasterGRN} can reduce refinement computation while bringing the generated image distribution closer to that of real COCO images.

\begin{figure*}[t]
    \centering
    \includegraphics[width=\textwidth]{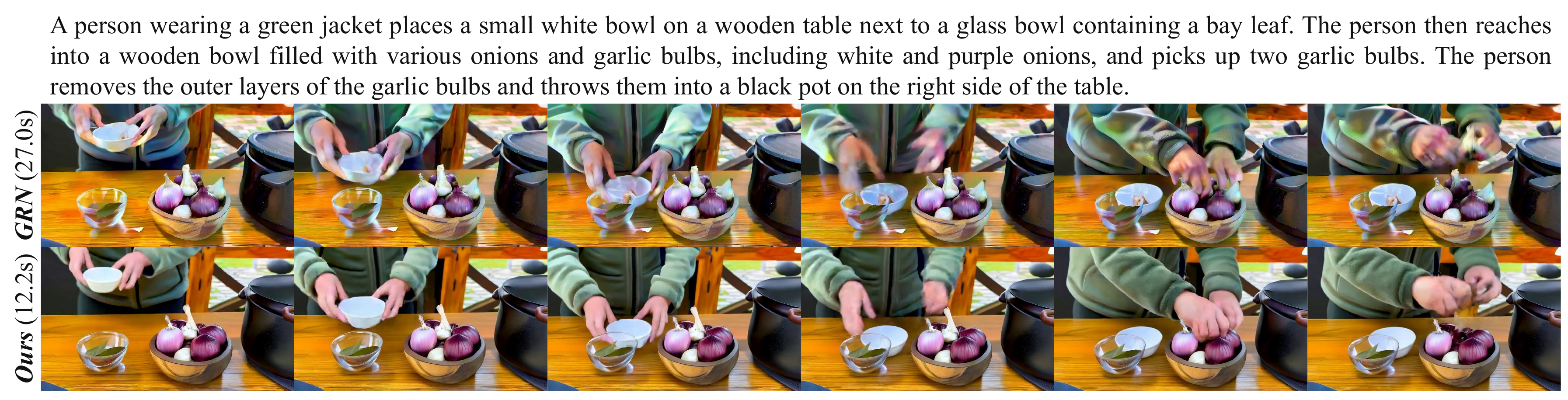}
    \vspace{-6mm}
    \caption{
        Qualitative comparison of text-to-video generation between GRN and \textbf{FasterGRN}. Six frames sampled at matching timestamps are shown for each method under the same text prompt and random seed.
    }
    \vspace{-3mm}
    \label{fig:video_comparison}
\end{figure*}

\vspace{-4mm}
\paragraph{Additional Results on Video Generation.}
As additional qualitative results, we apply \textbf{FasterGRN} to GRN-T2V-2B for video generation using the same prompt and seed as the original GRN.
Across the 6 frames shown in Fig.~\ref{fig:video_comparison}, both methods depict a person in a green jacket handling garlic at a wooden table.
In this example, \textbf{FasterGRN} reduces generation time from 27.0 to 12.2 seconds ($2.21\times$ speedup) while preserving the main subject and action.

\begin{table}[t]
\centering
\caption{
Quantitative comparison on
COCO2014 and COCO2017.
}
    \vspace{-1mm}
\label{tab:coco_comparison}
\setlength{\tabcolsep}{4pt}
\renewcommand{\arraystretch}{1.1}
\resizebox{\columnwidth}{!}{
\begin{tabular}{l|c|ccc|ccc}
\toprule
\multirow{2}{*}{Methods}
& \multirow{2}{*}{Speed$\uparrow$}
& \multicolumn{3}{c|}{COCO2014-30K}
& \multicolumn{3}{c}{COCO2017-5K} \\
\cline{3-8}
& &
FID$\downarrow$ & KID$\times10^{2}\downarrow$ & IS$\uparrow$
& FID$\downarrow$ & KID$\times10^{2}\downarrow$ & IS$\uparrow$ \\
\midrule
Baseline
& $1.0\times$
& 23.49 & 0.939 & 38.45
& 33.59 & 0.987 & 33.95 \\
\textbf{FasterGRN}
& $2.03\times$
& 22.19 & 0.924 & 39.15
& 32.16 & 0.955 & 34.89 \\
\bottomrule
\end{tabular}
}
\end{table}

\subsection{Ablation Study}
\label{sec:ablation_analysis}

\paragraph{Ablate the Stage Boundary.}
Table~\ref{tab:stage_transition} varies $t_s$ from 16 to 20 with all remaining hyperparameters fixed. Delaying starting step of \textit{detail refinement stage} reduces the speedup from $2.08\times$ to $1.88\times$, while CLIP, ImageReward, and HPSv2.1 consistently improve.
This indicates that retaining more original refinement trajectory slightly benefits semantic alignment and predicted human preference, but requires more computation.
GenEval and DPG scores do not improve monotonically, showing that a later transition does not uniformly benefit compositional generation.

\vspace{-2mm}
\paragraph{Ablate Each Acceleration Component.}
Table~\ref{tab:mechanism_ablation} examines the contributions of each component. Using only the conditional branch during the \textit{detail refinement stage} reduces guidance overhead but introduces some quality degradation. In contrast, \textit{ARE} accelerates generation while improving both benchmark scores over GRN, suggesting that adaptively advancing retention can reduce unnecessary refinement without compromising quality. \textit{STF} alone also accelerates generation while preserving comparable performance. Combining all three components yields the largest speedup of $2.03\times$ while largely preserving generation quality. These results highlight their complementary roles: conditional-only computation reduces guidance costs, while \textit{ARE} and \textit{STF} exploit redundancy across refinement iterations and token locations, respectively.

\begin{table}[t]
\centering
\caption{
Effect of the stage transition step $t_s$.
}
    \vspace{-1mm}
\label{tab:stage_transition}
\setlength{\tabcolsep}{4pt}
\renewcommand{\arraystretch}{1.08}
\resizebox{0.99\columnwidth}{!}{
\begin{tabular}{lcccccc}
\toprule
Method
& Speedup$\uparrow$
& GenEval$\uparrow$
& DPG$\uparrow$
& CLIP$\uparrow$
& IR$\uparrow$
& HPSv2.1$\uparrow$ \\
\midrule
GRN
& $1.00\times$
& 0.719 & 85.64 & 29.17 & 0.97 & 29.68 \\
\midrule
\textbf{FasterGRN} ($t_s=16$)
& $2.08\times$
& 0.713 & 84.49 & 28.94 & 0.90 & 28.72 \\
\textbf{FasterGRN} ($t_s=17$)
& $2.03\times$
& 0.711 & 84.73 & 29.08 & 0.91 & 28.81 \\
\textbf{FasterGRN} ($t_s=18$)
& $1.97\times$
& 0.711 & 84.52 & 29.10 & 0.92 & 28.87 \\
\textbf{FasterGRN} ($t_s=19$)
& $1.92\times$
& 0.709 & 84.55 & 29.13 & 0.93 & 28.93 \\
\textbf{FasterGRN} ($t_s=20$)
& $1.88\times$
& 0.706 & 84.88 & 29.15 & 0.93 & 29.00 \\
\bottomrule
\end{tabular}
}
\end{table}
 
\begin{table}[t]
\centering
\caption{
Ablate CFG-off, \textit{ARE}, and \textit{STF}.
}
    \vspace{-1mm}
\label{tab:mechanism_ablation}
\setlength{\tabcolsep}{4pt}
\renewcommand{\arraystretch}{1.08}
\resizebox{0.99\columnwidth}{!}{
\begin{tabular}{lcccccc}
\toprule
Method
& Speedup$\uparrow$
& GenEval$\uparrow$
& DPG$\uparrow$
& CLIP$\uparrow$
& IR$\uparrow$
& HPSv2.1$\uparrow$ \\
\midrule
GRN
& $1.00\times$
& 0.719 & 85.64 & 29.17 & 0.97 & 29.68 \\
\midrule
+CFG off
& $1.51\times$
& 0.705 & 83.77 & 28.91 & 0.85 & 28.48 \\
+\textit{ARE}
& $1.50\times$
& 0.726 & 85.89 & 29.26 & 1.00 & 29.50 \\
+\textit{STF}
& $1.15\times$
& 0.724 & 85.95 & 29.32 & 0.98 & 29.46 \\
\midrule
\textbf{FasterGRN}
& $2.03\times$
& 0.711 & 84.73 & 29.08 & 0.91 & 28.81 \\
\bottomrule
\end{tabular}
}
\end{table}

\section{Conclusion}
\label{sec:conclusion}

In this work, we analyze the generation process of GRN.
We first observe that semantic content is largely established early and tends to stabilize in later steps. We characterize the GRN generation process as consisting of the \textit{semantic establishment stage} and the \textit{detail refinement stage}.
Based on these observations, we introduce \textbf{FasterGRN}, a training-free framework to accelerate GRN generation.
It contains two novel speed-up modules, \textit{Adaptive Retention Expansion} and \textit{Selective Token Freezing}. 
They use predictive entropy plus bit flip rates to reduce step redundancy and token redundancy, while preserving frozen predictions during noise mixing. 
Experiments on diverse benchmarks demonstrate the noticeable speedup while largely preserving performance on GenEval and DPG-Bench and improving evaluation scores on COCO. 
Comparisons with existing methods further support the effectiveness of the framework. 
\clearpage

{
    \small
    \bibliographystyle{ieeenat_fullname}
    \bibliography{main}
}

\end{document}